\documentclass{article}
\usepackage{ijcai22}

\usepackage{times}

\usepackage{soul}
\usepackage{url}
\usepackage[hidelinks]{hyperref}
\usepackage[utf8]{inputenc}
\usepackage[small]{caption}
\usepackage{graphicx}
\usepackage{amsmath}
\usepackage{booktabs}
\author{Moshe Y. Vardi \and Zhiwei Zhang \thanks{Corresponding author: Zhiwei Zhang. }
\affiliations
Rice University, Houston, TX, USA 
\emails
\{vardi, zhiwei\}@rice.edu
}

\usepackage[table,xcdraw]{xcolor}
\usepackage{amsfonts}
\usepackage[misc]{ifsym}
\usepackage{amsmath}
\usepackage{booktabs}
\usepackage{amsthm}
\usepackage{multirow}
\usepackage{mathtools}
\usepackage{thmtools}
\usepackage{enumitem}
\usepackage{chngpage}
\usepackage[ruled,linesnumbered,noend]{algorithm2e}
\usepackage{lipsum,graphicx,subcaption}

\newtheorem{definition}{Definition}
\newtheorem{theorem}{Theorem}

\newtheorem{openquestion}{Open Question}

\newtheorem{proposition}{Proposition}
\newtheorem{remark}{Remark}
\newtheorem{problem}{Problem}

\newtheorem*{proof*}{Proof}
\newtheorem*{pfs*}{Proof Sketch}
\usepackage[noend]{algorithmic}

\newcommand{\addcite}[1]{\textcolor{blue}{[citation]}}

\usepackage[switch]{lineno}
\usepackage{xr}
\usepackage{subfiles}
\usepackage{physics}

\AtEndDocument{\refstepcounter{theorem}\label{finalthm}}
\AtEndDocument{\refstepcounter{proposition}\label{finalprop}}
\AtEndDocument{\refstepcounter{lemma}\label{finallemma}}

\title{Solving Quantum-Inspired Perfect Matching Problems via Tutte's Theorem-Based Hybrid Boolean Constraints \thanks{The author list has been sorted alphabetically by last name; this
should not be used to determine the extent of authors’ contributions.}}

\begin{document}

\maketitle

\begin{abstract}
Determining the satisfiability of Boolean constraint-satisfaction problems with different types of constraints, that is \emph{hybrid} constraints,  is a well-studied problem with important applications. We study a new application of hybrid Boolean constraints, which arises in quantum computing. The problem relates to constrained perfect matching in edge-colored graphs. 
While general-purpose hybrid constraint solvers can be powerful, we show that direct encodings of the constrained-matching problem as hybrid constraints scale poorly and special techniques are still needed. We propose a novel encoding based on Tutte's Theorem in graph theory as well as optimization techniques. Empirical results demonstrate that our encoding, in suitable languages with advanced SAT solvers, scales significantly better than a number of competing approaches on constrained-matching benchmarks. Our study identifies the necessity of designing problem-specific encodings when applying powerful general-purpose constraint solvers. 
\end{abstract}

\section{Introduction}
Constraint-satisfaction problems (CSPs) \cite{brailsford1999constraint} play a  fundamental role in mathematics, physics, and computer science. The Boolean SATisfiability problem (SAT) \cite{biere2009handbook} is a special class of CSPs, where each variable takes value from \{\texttt{True}, \texttt{False}\}. Solving SAT efficiently is of utmost significance in computer science, both theoretically and practically \cite{Vardi14a}. Though most effort of the SAT community has been put into solving conjunctive-normal-form (CNF) constraints, Boolean satisfiability that also allows hybrid (non-CNF) constraints is used in numerous applications in various areas, e.g., XOR constraints in cryptography  \cite{Biclique-Cryptanalysis-of-the-Full-AES}, pseudo-Boolean constraints in discrete optimization \cite{Graph-coloring-with-cardinality-constraints,nae-coloring,kyrillidis2020fouriersat}, and more. The combination of different types of constraints enhances the expressive power of Boolean formulas \cite{MLD}. 

Extensive research has been done for general-purpose hybrid Boolean constraints solving by a variety of techniques. Given the remarkable success of modern CNF-SAT solvers \cite{SAT-survey,minisat,maplesat,froleyks2021sat}, a natural approach is to encode hybrid constraints into CNF. Numerous types of encodings with different properties have been developed \cite{prestwich2009cnf}. Some high-level logic-modeling languages such as Answer Set Programming (ASP) \cite{lifschitz2019answer,gebser2019multi} are also based on CNF encoding. A different technique for hybrid constraints is to handle certain types of constraints natively by devising specialized solvers, such as CNF + XOR \cite{cmspaper} and pseudo-Boolean solvers \cite{cuttingToTheCore,openwbo}.  In addition, there also exist attempts for handling different types of hybrid constraints uniformly \cite{kyrillidis2020fouriersat,kyrillidis2021solving,kyrillidis2021continuous,kyrillidis2022dpms}.

 In this paper, we study a new application of hybrid constraints that arises in quantum computing, through quantum-inspired graph theoretical problems related to perfect matching with vertex color constraints. 
 Solving this problem efficiently provides essential insights into how to design quantum experiments \cite{jooya2016graph,duncan2020graph,krenn2022artificial} and advance the realization of large-scale quantum computers \cite{pan2012multiphoton} (see Section \ref{section:quantumInterpretation}).  
 Extant algorithmic work on this problem is far from developed, providing no scalable tool to the quantum-computing community. 
Tackling this problem by solving hybrid constraints is promising for several reasons. First, modern constraint solvers have been well-developed to solve large-scale instances. Second, additional properties of graph instances such as symmetry can be easily integrated as extra constraints, which can be difficult for pure graph theoretical algorithms. 

Applying general-purpose hybrid-constraint techniques to the problem above is, however, not trivial. A straightforward encoding of our constrained-matching problem leads to a blow-up of the length of the formula, as also observed in \cite{cervera2021design}. A more advanced encoding based on \emph{exact-one} constraints yields a considerably long quantified Boolean formula (QBF) with poor scalability. To address those issues, as the main technical contribution of this work, we propose an encoding for the  constrained-matching problem based on Tutte's Theorem in graph theory \cite{anderson1971perfect}. Tutte-Berge formula, as a generalized result of Tutte's Theorem, as well as related decompositions, are widely used in computer science and graph theory \cite{schrijver2003combinatorial}. The efficiency of our encoding originates from the short proof of the non-existence of perfect matchings that this encoding offers.   

In the experiment section, we evaluated our approach by solving novel benchmark problems that represent important tasks in quantum computing. For a comprehensive comparison, we also implemented a number of competing methods. We demonstrate that our Tutte encoding does not scale well without further optimizations and suitable languages. Specifically, expressing our encodings in high-level languages, such as Answer Set Programming (ASP) or pseudo-Boolean programming (PB), is more convenient. The resulting performance is, however, less than satisfactory. We discover that in conjunctive normal form (CNF) with additional optimizations and advanced SAT solvers, our encodings exponentially outperform other constraint-based and graph theoretical methods, significantly increasing the scalability of the state-of-the-art. Our implementations provide helpful tools for testing quantum properties efficiently. We conclude that in spite of the progress in general-purpose hybrid constraint solving, specialized encoding techniques are still required to obtain a scalable approach for specific problems.  

\section{Preliminaries} This section includes a background in hybrid Boolean SAT and definitions in graph theory which are needed for defining the quantum-inspired problem related to perfect matching.  
\label{sec:pre}

\begin{definition} {\rm(Boolean satisfiability and hybrid constraints)}
    Let $x=(x_1,...,x_n)$ be a sequence of $n$ Boolean variables. A Boolean constraint $f(x)$ is a mapping from a Boolean vector $\{\texttt{True,False}\}^n$ to $\{\texttt{True},\texttt{False}\}$. A formula $f=c_1\wedge c_2\wedge \dots \wedge c_m$ is the conjunction of Boolean constraints $\{c_i\}_{i=1,\cdots,n}$. There can be multiple types of constraints, e.g.:
	\begin{itemize}
	    \item Disjunctive Constraints, e.g.,  $(x_1\vee x_2 \vee x_3)$. 
	    \item Cardinality constraint, e.g., $x_1+x_2+x_3\ge 2$, where \texttt{False} and \texttt{True} are interpreted as $0$ and $1$ respectively. If the constraint is in the form of $\sum_{x\in X}x=k$, then it can be written as $\texttt{Exact-k}(X)$.
	    \item Pseudo-Boolean (linear) constraints, e.g., $3x_1-4x_2+5x_3\ge 0$.
     \item XOR constraints, e.g., $x_1\oplus x_2\oplus x_3$.
	\end{itemize}
	 A set of constraints is \emph{satisfiable} if there exists a Boolean assignment (solution) that evaluates all constraints to \texttt{True}.
\end{definition}

\begin{figure}[t]
\centering
\includegraphics[scale=0.16]{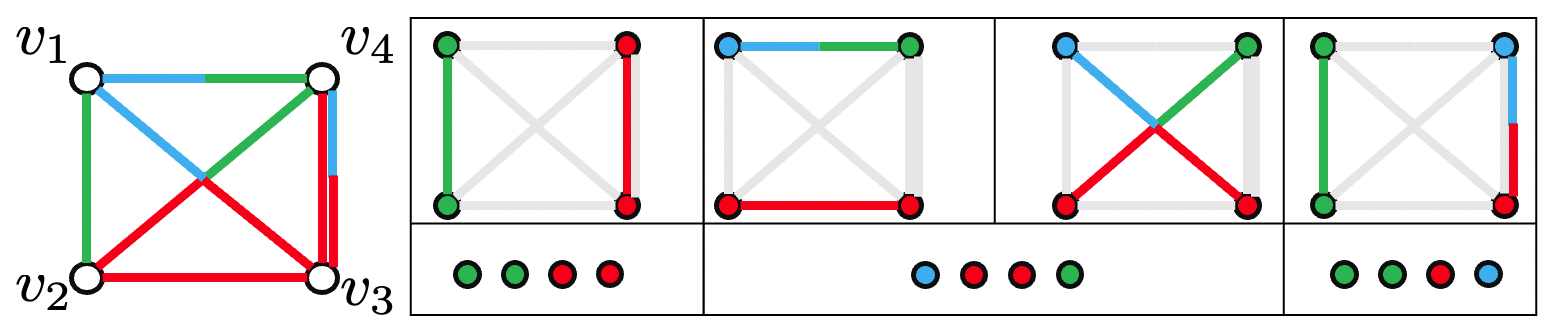}
\caption{Left: a graph $G$ with bicolored edges. $v_3$ and $v_4$ are connected by two edges with different colors. Right: all perfect matchings of $G$ and their inherited vertex colorings \protect\cite{vardi2022quantum}.}
\label{fig:graph}
\end{figure}

Next, we introduce concepts that relate to the quantum-inspired graph theoretical problem.
\begin{definition}
{\rm (Bicolored Graphs \cite{gu2019quantum3,vardi2022quantum})} In this paper, a ``graph'' refers to an undirected, non-simple \footnote{Self-loops are prohibited but multiple edges between the same pair of vertices are allowed.} graphs. A bicolored graph $G$ is a tuple $(V,E,d)$, where $V$ is the set of vertices with $|V|$, $E$ is the set of bicolored edges, and $d$ is the number of colors. For each edge $e\in E$ that connects $u$ and $v$, there are two colors $c_u^e,c_v^e\in \{1,\cdots, d\}$ that are associated with $u$ and $v$, respectively, i.e., $e=\{(u,c_u^e),(v,c_v^e)\}$. If $c_u^e=c_v^e$, we call the edge $e$ monochromatic, otherwise $e$ is bicolored. See Figure \ref{fig:graph} (left) as an example.
 \end{definition} 

\begin{definition}
{\rm (Vertex coloring)} A vertex coloring of graph $G=(V,E,d)$ is a mapping $c$ from vertices to colors, i.e., $c:V\to \{1,\cdots,d\}$. Alternatively one can write  $c\in {\{1,\cdots,d\}}^V$. For a coloring $c$ and a color $i\in \{1,\cdots,d\}$, we use \texttt{count}$(c,i)$ to denote the number of appearances of color $i$ in $c$.
\end{definition} 

\begin{definition}
{\rm (Perfect matchings and their inherited vertex colorings) \cite{krenn2019questions}} A perfect matching $P$ of a graph $G$ is a subset of edges, i.e., $P\subseteq E$, such that for each vertex $v\in V$, exactly one edge in $P$ is adjacent to $v$. For a perfect matching $P$ of a bicolored graph $G$, its \emph{inherited vertex coloring}, denoted by $c^P:V\to \{1,\cdots, d\}$ is defined as follows: for each vertex $v$, let $e$ be the unique edge that is adjacent to $v$, then $c^P(v)=c_v^e$. See Figure \ref{fig:graph} for examples.
\end{definition} 

\section{Quantum-Inspired Perfect-Matching-Related Problems}
\label{sec:quantum}
In this section, we introduce the graph-theoretical problem that relates to vertex-color-constrained perfect matching, inspired by quantum computing. Roughly speaking, the problem asks whether for \emph{each}  vertex coloring in a set of ``legal" vertex colorings, a bicolored graph has at least one perfect matching with this inherited vertex coloring. 

\begin{problem}
{\rm (Perfect matchings exist for all legal vertex colorings, abbrv. FORALL-PMVC)} Consider a bicolored graph $G$ and a set of $\mathcal{C}\subseteq \{1,\cdots, d\}^V$ ``legal" vertex colorings. Does $G$ satisfy that, for each vertex coloring $c\in \mathcal{C}$, $G$ has at least one perfect matching with inherited vertex coloring $c$?
\label{prob:mainProblem}
\end{problem} 

The hardness of FORALL-PMVC depends heavily on the set of legal vertex colorings $\mathcal{C}$. $\mathcal{C}$ can be defined by enumerating all members of the set or by constraints. Below, we list several legal sets of vertex colorings with interest in quantum computing \cite{gu2019quantum3,vardi2022quantum}.

\begin{figure}[t]
	\centering
\includegraphics[scale=0.12]{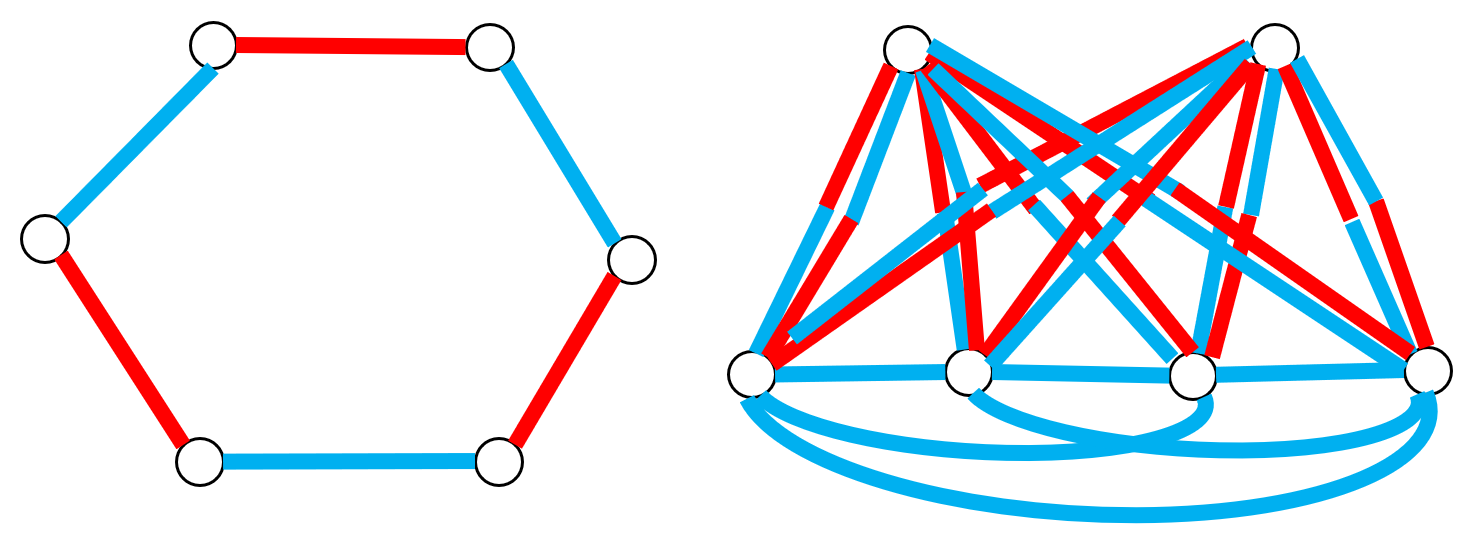}
\caption{Left: A cycle satisfies the FORALL-PMVC condition of \texttt{GHZ}(6,2). The graph contains perfect matchings for both of two monochromatic vertex colorings. Right: A graph that satisfies the FORALL-PMVC condition for \texttt{Dicke}(6,2). The graph contains perfect matchings for all vertex colorings that have two red vertices.}
\label{fig:graphExamples}
\end{figure}

    \begin{itemize}
        \item GHZ State: the legal colorings are monochromatic, i.e., $\mathcal{C}=\{(1,1,\cdots,1),(2,2\cdots,2),\cdots, (d,d,\cdots,d)\}$, denoted as \texttt{GHZ}$(|V|,d) = {1}/{\sqrt{d}}\cdot\sum_{i=1}^{d}\ket{i}^{\oplus |V|}$.
        \item W State: the graph has two colors ($d=2$). $\mathcal{C}=\{c\in \{1,\cdots, d\}^V|\texttt{count}(c,1)=1\}$, i.e, there must be exactly one vertex with color $1$ and $|V|-1$ vertices with color $2$, denoted as $\texttt{W}(|V|)={1}/{\sqrt{|V|}}\cdot\hat{S}(\ket{2} ^{\oplus (|V|-1)}\ket{1} ^{\oplus 1})$
        \item Dicke State: a generalization of W State. $\mathcal{C}=\{c\in \{1,\cdots, d\}^V|\texttt{count}(c,1)=k\}$ ($d=2$), denoted as $\texttt{Dicke}(|V|,k)={1}/{\sqrt{\binom{|V|}{k}}}\cdot\hat{S}(\ket{2} ^{\oplus (|V|-k)}\ket{1} ^{\oplus k})$.
    \end{itemize}
Complete bicolored graphs trivially satisfy the FORALL-PMVC condition for an arbitrary legal coloring set. Nevertheless, graphs with interest in quantum computing are those satisfying the FORALL-PMVC condition with as few edges as possible. In Figure \ref{fig:graphExamples}, we show examples of such graphs.

The problem of checking the existence of perfect matching is well-known to be polynomial, via the Blossom algorithm \cite{edmonds1965paths}. We now show that if the size of legal colorings, i.e., $|\mathcal{C}|$, is polynomially bounded, then FORALL-PMVC is tractable. The idea is that one can enumerate all legal colorings and verify the existence of perfect matching individually, as formalized in the following proposition.

\begin{definition} {\rm (Induced graph w.r.t. a vertex coloring)}
Given a bicolored graph $G=(V,E,d)$ and a vertex coloring $c$, let $G_c=(V,E_c,d)$ be the subgraph where $E_c$ is a subset of $E$, defined as follows:
$$
E_c=\{e|e=\{(u,c_u^e),(v,c_v^e)\}\in E, c_u^e = c(u),c_v^e=c(v)\}.
$$
In other words., in $G_c$, we keep only the edges whose colors agree with the vertex coloring c. 
\end{definition}

\begin{proposition}
    A bicolored graph $G$ has a perfect matching with inherited vertex coloring $c$ iff $G_c$ has a perfect matching. 
    \label{prop:reducedGraph}
\end{proposition}
With Proposition \ref{prop:reducedGraph}, we can design a polynomial algorithm for FORALL-PMVC with polynomially many legal colorings, as shown in Algorithm \ref{algorithm:enum-blossom}. We assume \texttt{Blossom}$(G)$ is an algorithm that returns \texttt{True} iff. $G$ has a perfect matching.

\begin{algorithm}[ht]
    \SetAlgoLined
    \SetKwInOut{Input}{Input}
    \SetKwInOut{Output}{Output}
    \Input{Bicolored graph $G$, set of legal colorings $\mathcal{C}$.}
    \Output{Whether $G$ satisfies the FORALL-PMVC condition w.r.t. $\mathcal{C}$.}
    \vspace{0.1cm}
    \hrule
    \vspace{0.1cm}
    \For{$c\in \mathcal{C}$}{
    Let $G_c$ be the induced graph of $G$ w.r.t. $c$.\\
    \lIf{\texttt{Blossom}($G_C$)==\texttt{False}}{\Return \texttt{False}}}
    \Return \texttt{True}\\
 \caption{\texttt{Enum-Blossom}: An Enumeration-Based Algorithm for FORALL-PMVC}
 \label{algorithm:enum-blossom}
\end{algorithm}
Algorithm \ref{algorithm:enum-blossom} runs in $O(|\mathcal{C}|\cdot |E|\cdot|V|^2)$. Therefore it provides an efficient algorithm for the FORALL-PMVC problem for GHZ State ($|\mathcal{C}|=d$) and W State ($|\mathcal{C}|=|V|$). The challenging part of FORALL-PMVC is when $|\mathcal{C}|$ is exponential, such that it is infeasible to enumerate $\mathcal{C}$. As an example, \texttt{Dicke}$(n,n/2)$ has $\binom{n}{n/2}=\Theta(2^n/\sqrt{n})$ legal colorings.

The lower bound of the computational complexity of FORALL-PMVC for exponentially many legal colorings remains open, though in \cite{vardi2022quantum} the complexity of a related problem is addressed for some special cases. It is not obvious how to reduce a well-known NP-hard or W[1]-hard problem to FORALL-PMVC. 

\begin{openquestion}
    Is FORALL-PMVC of Dicke State NP-hard?
\end{openquestion}

\begin{openquestion} Is FORALL-PMVC of \texttt{Dicke}$(n,k)$ fixed-parameter tractable (FPT) or W[1]-hard w.r.t. $k$? \footnote{In \cite{gaspers2012don}, it is proven that deciding whether an uncolored graph has a Tutte set of exactly $k$ vertices is W[1]-hard.}
\end{openquestion}

As for the upper bound, the problem is in co-NP, provided that checking if a coloring $c$ is in $\mathcal{C}$ is in PTIME.
\begin{proposition}
   \label{prop:coNP}
    FORALL-PMVC is in co-NP.
\end{proposition}

\begin{proof*}
    If a graph does not satisfy the condition of FORALL-PMVC, then a witness can be obtained as a legal coloring $c\in \mathcal{C}$ such that $G_c$ has no perfect matching. Such a witness $c$ can be verified in polynomial time by the Blossom Algorithm.    
\end{proof*}
    
If one views a coloring $c$ as a witness, then the algorithm for checking the witness can be relatively sophisticated, such as the Blossom Algorithm. There exists, however, a more powerful form of witness that allows a simpler checking algorithm, which leads to our approach for solving FORALL-PMVC, as shown in Section \ref{section:BooleanEncodings}.

\begin{figure*}[ht]
\centering
\includegraphics[scale=0.22]{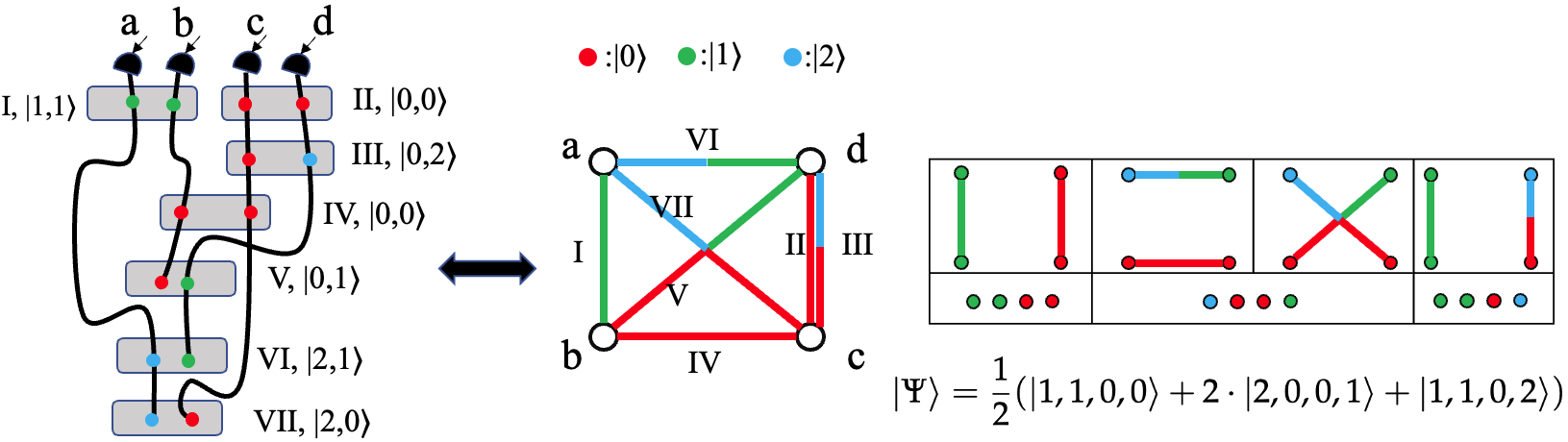}
\caption{Left: a quantum optical circuit corresponding to the graph in Figure \ref{fig:graph}. Each crystal (shaded box) corresponds to an edge in the graph. Each optical path is converted into a vertex in the graph. The modes of photons translate to edge colors. Right: all PMs in the graph reflect the quantum state of the optical circuit as a superposition of coincidences. \protect\cite{vardi2022quantum}}
\label{fig:circuitPM}
\end{figure*}

\section{Quantum Motivation and Impact of FORALL-PMVC}
\label{section:quantumInterpretation}
\subsection{Perfect Matchings and Quantum Circuits}
The motivation of PMVC from the perspective of quantum-experiment design was originally introduced in \cite{vardi2022quantum}, which we recall here to make this paper self-contained. An example of a (simplified) \emph{quantum optical circuit} is shown in Figure \ref{fig:circuitPM} (left). A non-linear \emph{crystal} (shaded box in Figure \ref{fig:circuitPM}) can emit a pair of entangled \emph{photons} simultaneously if activated by a laser-pump power. Each photon has a \emph{mode}, encoded by an integer index. Photons emitted by crystals travel on  \emph{optical paths}. A \emph{receiver}, which detects the arrival of a photon and identifies photon mode, is placed at the end of each optical path. Each crystal is associated with two optical paths \cite{gu2019quantum2}. 

A \emph{coincidence} of a quantum optical circuit happens when each receiver detects exactly one photon, due to the activation of some crystals. The \emph{quantum state} of a coincidence is the mode of photons caught by all receivers in the coincidence. The quantum state of an optical circuit is the \emph{superposition} \cite{ballentine2014quantum} of the quantum states of all coincidences. 

The optical circuit in Figure \ref{fig:circuitPM} (left) has $7$ nonlinear crystals. A coincidence happens only when crystals in \{I,II\}, or \{I,III\}, or \{IV,VI\} or \{V,VII\} are activated simultaneously.

In \cite{gu2019quantum2}, a coincidence of an optical circuit is shown to be equivalent to a perfect matching under vertex-color constraints in an undirected graph $G$ with bi-colored edges, as described below. 
\begin{enumerate}
    \item Each optical path corresponds to a vertex in $G$.
    \item Each crystal corresponds to an edge in $G$. This edge connects two vertices corresponding to two optical paths to which the crystal can emit photons.
    \item The modes of photons emitted by crystals correspond to edge colors. Since a crystal can emit two photons with different modes, edges in $G$ are bi-colored.
    \item A coincidence of an optical circuit corresponds to a perfect matching $P$ in $G$. The set of edges in $P$ indicates activated crystals. The quantum state of the coincidence corresponds to the inherited vertex coloring of $P$.
    \item Each crystal has an amplitude as a complex number, corresponding to the edge weight in the graph. The amplitude of a coincidence is the product of the amplitudes of all activated crystals. The weight of a perfect matching is the product of the weights of all its edges. 
\end{enumerate}

\subsection{Impact of FORALL-PMVC}
It is shown in a series of work \cite{gu2019quantum3,gu2019quantum2} that designing quantum circuits can be done by constructing a graph with bi-colored edges. The ultimate goal of this line of research is to automatically construct bi-colored graphs by combinatorial approaches for arbitrary quantum states, aiming to contribute to the development of large-scale and reliable quantum computers. 

The graphs proposed originally in \cite{gu2019quantum2} allow complex edge weights, representing the amplitude and phase of the optical components. A quantum state of an optical circuit (resp., graph) can be viewed as a superposition of coincidences (resp., PMs) with predefined ``legal''  quantum states. An optical circuit (resp., a graph) exhibits a quantum state if 1) the total amplitude (resp., weight) of all coincidences (resp., PMs) of each legal state is $1$, and 2) all coincidences (resp., PMs) with an ``illegal'' state have total amplitude (resp., weight) $0$ \cite{krenn2017quantum}.

The original graph construction problem displays a continuous nature, while in this paper (FORALL-PMVC) as well as \cite{vardi2022quantum} (EXISTS-PMVC) we study two simplified, discrete problems, which can be viewed as focusing on the case where edge weights are real and positive. The simplified, discrete setting is not only important in graph theory but also of interest in quantum experiments. Graphs with real, positive edge weights correspond to circuits with optical components with zero-phase. Such a circuit is efficient to set up.   Although the existence of graphs with real, positive weights of GHZ states has been studied \cite{krenn2019questions}, the simplified setting still remains open and challenging for many other quantum states, e.g., Dicke State. A key feature of our approach is its generality in terms of algorithmically accepting arbitrary user-defined quantum states. In the future, we aim to extend our approach to allow complex weights.

FORALL-PMVC and EXISTS-PMVC test two conditions of whether a graph can be a candidate for displaying a quantum state, under the real-positive-weight-assumption. The algorithms for solving those two decision problems can be applied as building blocks for further developing the actual graph constructors. For instance, a search-based graph constructor can use those algorithms as oracles for quickly pruning unwanted structures in the space of all bi-colored graphs. 

\section{Boolean Encodings for FORALL-PMVC} 
\label{section:BooleanEncodings}
In this section we introduce our hybrid-constraint-based approach for FORALL-PMVC. We show that a natural and straightforward \texttt{Exact-One} encoding yields a non-scalable QBF approach. Rather, we exploit the powerful witness of the non-existence of perfect matching provided by Tutte's Theorem to design a novel encoding. We further propose optimization techniques for our Tutte's Theorem-based encoding to enhance the  performance.

\subsection{The Exact-One Encoding for PM and QBF Encoding for FORALL-PMVC} \label{subsection:exactone} From the definition of perfect matching, it is natural to use \texttt{ExactOne} constraints to encode perfect-matching-related problems. For each $e\in E$, we introduce a variable \texttt{edge}$_e$, which is \texttt{True} iff $e$ is in the perfect matching:
$$
\texttt{PM}=\{\texttt{ExactOne}(\{\texttt{edge}_e\}_{e\in \texttt{adj}(v)})\}_{v\in V}.
$$

\texttt{PM} is satisfiable iff the graph has a perfect matching. The variables $\{\texttt{edge}_e\}_{e\in E}$ in \texttt{PM} are existentially quantified. As FORALL-PMVC requires that for each legal coloring the graph has a perfect matching, one can add a universally quantified layer corresponding to vertex colorings. For each vertex $v$ and color $i$, we define a variable $\texttt{vc}_{v}^i$, which is \texttt{True} iff. vertex $v$ has color $i$.  Then FORALL-PMVC can be expressed in the following $\forall$-$\exists$ quantified Boolean formula (QBF):
\begin{equation}
\begin{split}
&\forall\{\texttt{vc}_v^i\}_{v\in V, 1\le i\le d} \exists\{\texttt{edge}_e\}_{e\in E},\\
&(\texttt{ValidColoring}\wedge \texttt{LegalColoring})\to \texttt{PM}.
\label{eq:qbf}
\end{split}
\end{equation}

\texttt{ValidColoring} is satisfied if the assignment to the vertex-color variables $\{\texttt{vc}\}$ represents a valid vertex coloring, i.e. each vertex has exactly one color:
$$
\texttt{ValidColoring}= \{\texttt{ExactOne}(\{\texttt{vc}_v^i\}_{1\le i\le d})\}_{v\in V}.
$$
\texttt{LegalColoring} includes the constraints that define the set $\mathcal{C}$ of legal coloring. For example, for \texttt{Dicke}$(|V|,k)$, we have 
$$
\texttt{LegalColoring}=\{\sum_{v\in V}\texttt{vc}_v^1=k\}.
$$
Therefore the QBF formula (\ref{eq:qbf}) is satisfiable iff. the graph satisfies the FORALL-PMVC condition. Extant QBF solvers, however, accept only problems in CNF format, which requires encoding the implication and cardinality constraints in (\ref{eq:qbf}) to CNF. Even using scalable encoding techniques such as Tseytin encoding \cite{tseitin1983complexity}, the length of the CNF-QBF formula 
increases rapidly as the graph scales, 
which makes QBF solvers perform poorly, as demonstrated in Section \ref{section:experiments}.

\subsection{A Tutte's Theorem-Based Encoding}
\label{subsec:tutte}
We show that it is possible to encode FORALL-PMVC with only existential quantifiers, such that a broader set of solvers, e.g., SAT solvers and hybrid Boolean solvers, can be applied. It is enabled by the witness of the non-existence of perfect matching given by Tutte's Theorem.

\begin{theorem} {\rm (Tutte's Theorem)} Let the number of odd connected components (connected components having an odd number of vertices) of a graph $G$ be $\#odd(G)$. For a subset $V'\subseteq V$ of vertices, let $G[V']$ be the induced subgraph of $G$ on $V'$. A graph $G$ has no perfect matching if and only if there exists a set $S\subseteq V$ of vertices such that $$\#odd(G[V\setminus S])>|S|.$$ We call such a set $S$ a \emph{Tutte set}. 
\end{theorem}

 For example, consider the complete bipartite graph $K_{n-2,n}$, where every vertex of the first set containing $(n-2)$ vertices is connected to every vertex of the second set including $n$ vertices. It is easy to see that $K_{n-2,n}$ has no perfect matching. All $(n-2)$ vertices in the first set form a unique Tutte set, since removing those $n-2$ vertices yields $n$ isolated vertices, i.e., $n$ odd connected components.

The idea is, with Tutte's Theorem, we can convert the problem of ``finding a legal coloring whose induced graph has no perfect matching" to ``finding a legal coloring that allows a Tutte set''. If for all legal colorings, there exists no Tutte set, then the graph satisfies the FORALL-PMVC condition. In the following, we introduce the details of our encoding for FORALL-PMVC based on Tutte's Theorem.

For each vertex $v$, we introduce a variable $T_v$, which is \texttt{True} iff $v$ is in the Tutte set. 
For each edge $\{(u,a),(v,b)\}\in E$, we use a variable $\texttt{e}_{uv}^{ab}$ to represent whether this edge remains in the induced subgraph $G[V\setminus S]_c$ w.r.t. a vertex coloring $c$. Then we have:
\begin{equation}\nonumber
\begin{split}
     &\texttt{RemainingEdges}\\
     =& \{\texttt{e}_{uv}^{ab} \leftrightarrow  \big( \neg T_v\wedge \neg T_u \wedge \texttt{vc}_u^a \wedge \texttt{vc}_v^b \big)\}_{\{(u,a),(v,b)\}\in E}.
\end{split}
\end{equation}    
 
One of the technical challenges of encoding Tutte's Theorem is how to count odd connected components of $G[V\setminus S]_c$. For each vertex $v$ and an index $i$ of a connected component, 
we introduce a variable $\texttt{cc}_v^i$, which is \texttt{True} iff vertex $v$ is in the $i$-th connected component. Although there can be up to $|V|$ non-empty connected components in $G[V\setminus S]_c$ (consider $|V|$ isolated vertices), multiple connected components are often empty in practice. For two vertices $u$ and $v$, if they are connected by an edge that remains in $G[V\setminus S]_c$, then they must be in the same connected component, represented by the following constraints:
\begin{equation}\nonumber
\begin{split}
&\texttt{ConnectedComponent}\\
=&\{\texttt{e}_{uv}^{ab}\to \bigwedge_{1\le i\le |V|}(\texttt{cc}_{v}^i\leftrightarrow\texttt{cc}_{u}^i)\}_{\{(u,a),(v,b)\}\in E.}
\end{split}
\end{equation}
It is worth pointing out that \texttt{ConnectedComponent} does not enforce different components to be indexed differently. Nevertheless, \texttt{TutteCondition} below is in favor of using as many indices of connected components as possible.

For each vertex $v$ that is not in the Tutte set, we require $v$ to be in exactly one connected component. If $v$ is in the Tutte set, then it does not belong to  any of the connected components, represented by:
$$
\texttt{ValidComponent}=\{\texttt{ExactOne}(\{\texttt{cc}_v^i\}_{1\le i\le |V|}\cup \{T_v\})\}_{v\in V}
$$
 
For each index $i$ $(1\le i\le |V|)$ of connected components, we use a variable $\texttt{Odd}_i$ to indicate whether the connected component with index $i$ has an odd number of vertices. The set of variables $\{\texttt{Odd}\}_i$ complies with the following set of XOR constraints:
$$
\texttt{Odd} = \{\texttt{Odd}_i\leftrightarrow\bigoplus_{v\in V}\texttt{cc}_{v}^i\}_{1\le i\le |V|}.
$$
Finally, we enforce that the number of odd connected components in $G[V\setminus S]_c$ is greater than the size of the Tutte set by a cardinality constraint:
$$
\texttt{TutteCondition}=\{\sum_{i=1}^{|V|} \texttt{Odd}_i> \sum_{v\in V}T_v\}.
$$

Let \texttt{TutteEncoding} be the union of constraints in \texttt{RemainingEdges}, \texttt{ConnectedComponent}, \texttt{ValidComponent}, \texttt{Odd} and \texttt{TutteCondition}, as well as \texttt{ValidColoring} and \texttt{QuantumState} defined previously. Then we have the following proposition, whose proof is delayed to the appendix.

\begin{proposition} {\rm (Correctness of Tutte encoding)} \texttt{TutteEncoding} is  satisfiable iff the bicolored graph does not satisfy the FORALL-PMVC condition.
\label{prop:tutteencoding}
\end{proposition}

 Our Tutte encoding uses $|V|^2 + |E| +(d+2)\cdot |V|$ variables.

\begin{remark} The Tutte encoding can be alternatively written in ILP, which allows the Boolean vectors for expressing colorings and indices of connected components to be replaced by a single integer variable. Meanwhile, the constraints for \texttt{ValidComponent} and \texttt{ValidColoring} are no longer necessary in the ILP encoding. The evaluation of Tutte encoding on ILP solvers is left as future work. 
\end{remark}

\subsection{Optimizations of the Tutte Encoding}
\label{section:optimization}
The Tutte encoding we proposed above can be improved, by exploiting the symmetric nature of Tutte set and the input graph. In the following, we describe optimization techniques that reduce the search space by 1) using fewer variables and 2) eliminating redundant solutions. As will be seen in the experiment section, the optimized Tutte encoding scales exponentially better than the vanilla one proposed above. 

\subsubsection{4.3.1 Exploiting the symmetry in Tutte's Theorem} \label{para:opt} Each vertex in $V\setminus S$ belongs to a connected component. Given a solution of the Tutte encoding, permuting the indices of connected components also generates a solution. To reduce the solution space, we can leave each vertex with fewer options regarding the index of the connected component. Specifically, for a vertex $v$ with $1\le v\le |V|$, we only keep variables $\texttt{cc}_v^1,\cdots, \texttt{cc}_v^{v}$. That is, a vertex $v$ must belong to a connected component whose index is smaller or equal to $v$. $\frac{|V|(|V|-1)}{2}$ variables are removed from the vanilla Tutte encoding.

Additionally, we add constraints to reduce the number of solutions. In spite of that $G[V\setminus S]_c$ can have up to $|V|$ connected components, the number of non-empty connected components is often much smaller than $|V|$. Therefore, for the same Tutte set, there is still considerable flexibility in the assignment of connected component indices, which we aim to further reduce. If a vertex $v$ is in the Tutte set or belongs to a connected component with an index smaller than $v$, 
 we can rule out all solutions that contain non-empty connected components indexed by $v$, expressed by the following constraints:
\begin{equation}\nonumber
    \begin{split}
\texttt{OPT} =& \{T_v\to \big(\bigwedge_{v\le u\le |V|}\neg\texttt{cc}_u^v\big)\}_{v\in V}\\
              &\cup \{\big(\bigvee_{1\le i < v}\texttt{cc}_v^i \big) \to \big(\bigwedge_{v\le u\le |V|}\neg\texttt{cc}_u^v\big)\}_{v\in V}.
    \end{split}
\end{equation}

\subsubsection{4.3.2 Exploiting the symmetry of the input graph}\label{para:gs}
One of the advantages of constraint-based approaches is that the properties of the input can be easily taken into account as extra constraints. Previous work has been done in symmetry-breaking by leveraging graph automorphism when solving combinatorial problems by propositional logic \cite{aloul2006efficient}. Since adding all potential symmetry-breaking constraints often leads to an exponential overhead, in this work we apply a lightweight symmetry-breaking strategy. If there exists a set $U\subseteq V$, such that all vertices are equivalent in graph $G$, i.e., permuting vertices in $U$ leads to a graph isomorphism to $G$. Then we want the solution of the Tutte encoding to have the smallest lexicographically  representative in $U$. 
For such a set $U$ and an order of vertices in $U$: $(u_1,\cdots, u_{|U|})$, we add the following constraints:
$$
\texttt{GraphSymmetryBreaking} = \{T_{u_i}\to T_{u_{i-1}}\}_{2\le i\le |U|}.
$$

Note that our approach is not the only way of breaking symmetry. There exists a trade-off between the size of constraints and solution-space reduction \cite{walsh2006general}. We leave investigating the interaction of different symmetry-breaking techniques on FORALL-PMVC as a future work.

\section{Experiment Results}
\label{section:experiments}
This section evaluates our Tutte encoding on solving FORALL-PMVC problems.

\subsection{Implementations and Methods in Comparison}

\noindent\textbf{RQ1}. How do the optimization techniques in Section \ref{section:optimization} improve the performance of the Tutte encoding?\\
\textbf{RQ2}. Among conjunctive normal form (CNF), pseudo-Boolean+XOR (PBXOR), and Answer Set Programming (ASP), which language fits best for the Tutte encoding?\\
\textbf{RQ3}. How do methods based on Tutte encoding perform on 
 FORALL-PMVC benchmarks? 
\subsection{Implementations and Competing Methods}
We implemented the Tutte encoding in CNF formulas (TutteCNF), pseudo-Boolean+XOR constraints (TuttePBXOR), and Answer Set Programming (TutteASP). Additionally, we implemented the QBF-based approach in Section \ref{subsection:exactone} and Algorithm \ref{algorithm:enum-blossom} based on the Blossom Algorithm. Our implementations are based on the following software.
\begin{itemize}
\item For solving ASP programs, we use Clingo \cite{gebser2019multi}. Due to the powerful expressiveness of ASP, the implementation of Tutte encoding is less than 30 lines of code, which is considerably more compact than CNF and pseudo-Boolean encoding.
    \item  For solving CNF Tutte encoding, we use two SAT solvers: Kissat \cite{fleury2020cadical}, a winner of recent SAT Competitions, and Clasp \cite{gebser2007clasp}, the underneath SAT solver in Clingo in order for fairly comparing CNF with ASP. 
    \item For solving pseudo-Boolean+XOR encoding, we use LinPB \cite{yang2021engineering}, a recent advanced hybrid Boolean constraint solver with native XOR support. 
    \item We implemented Alg. \ref{algorithm:enum-blossom} (Enum-Blossom) with Python package \texttt{networkx} \cite{hagberg2020networkx}.
    \item For solving QBF formulas, we use DepQBF, a medal winner of QBF evaluations \cite{lonsing2010depqbf}.
\end{itemize} 
Besides the vanilla Tutte encoding in Section \ref{subsec:tutte}, we use Tutte[Language]+Opt to denote the Tutte encoding including constraints in \texttt{OPT} defined in Section \ref{para:opt}.1 and with unnecessary variables removed. We use Tutte[Language]+Opt+GS to denote the Tutte encoding including constraints in both \texttt{OPT} and \texttt{GraphSymmetryBreaking} in Section \ref{para:gs}.2. 

 All experiments were run on single CPU cores of a Linux cluster at 2.60-GHz and with 16 GB of
RAM. We set the time limit for each problem instance to $1000$ seconds. The implementations and benchmarks can be found in the GitHub repository: \url{https://github.com/zzwonder/PMVC}.

\paragraph{Experiment 1: Proving the Non-existence of PM of Uncolored Graphs.}
We first evaluate the efficiency of Tutte encoding in identifying the non-existence of PM of uncolored graphs, before solving FORALL-PMVC problems. The benchmarks include complete bipartite graphs $K_{n-2,n}$ for $n\in\{10,12,14,\cdots,70\}$. All graphs have no perfect matching. The first partition with $n-2$ vertices forms the unique Tutte set. We compare the Tutte encoding with the \texttt{Exact-One} encoding in Section \ref{subsection:exactone}, whose unsatisfiability indicates the non-existence of perfect matching.  We also include the Blossom Algorithm in comparison. The running time for each algorithm to prove $K_{48,50}$ has no perfect matching is shown in Table \ref{table:bi} and the comprehensive results are delayed to the appendix. Each running time is the median of $100$ independent executions.

\begin{figure}[t]
	\centering
\includegraphics[scale=0.46]{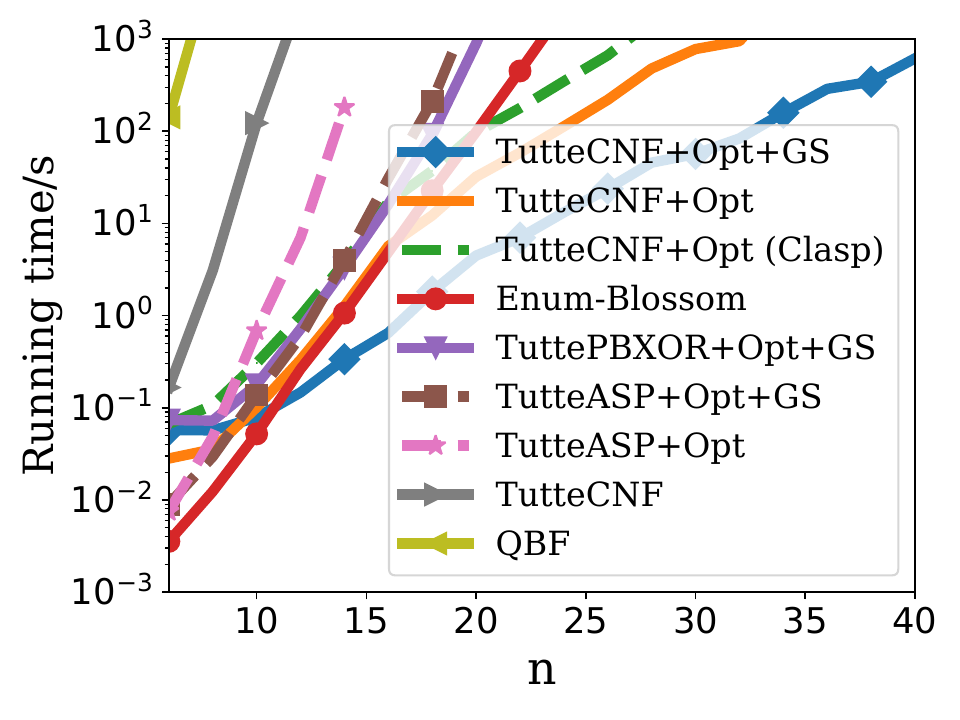}
\caption{Running time for verifying the FORALL-PMVC condition for graphs that satisfy \texttt{Dicke}$(n,n/2)$. The unsatisfactory performance of TutteASP and TuttePBXOR (without +Opt, +GS) is omitted for readability.}
\label{fig:dickenn2}
\end{figure}

\begin{figure}[t]
	\centering
\includegraphics[scale=0.46]{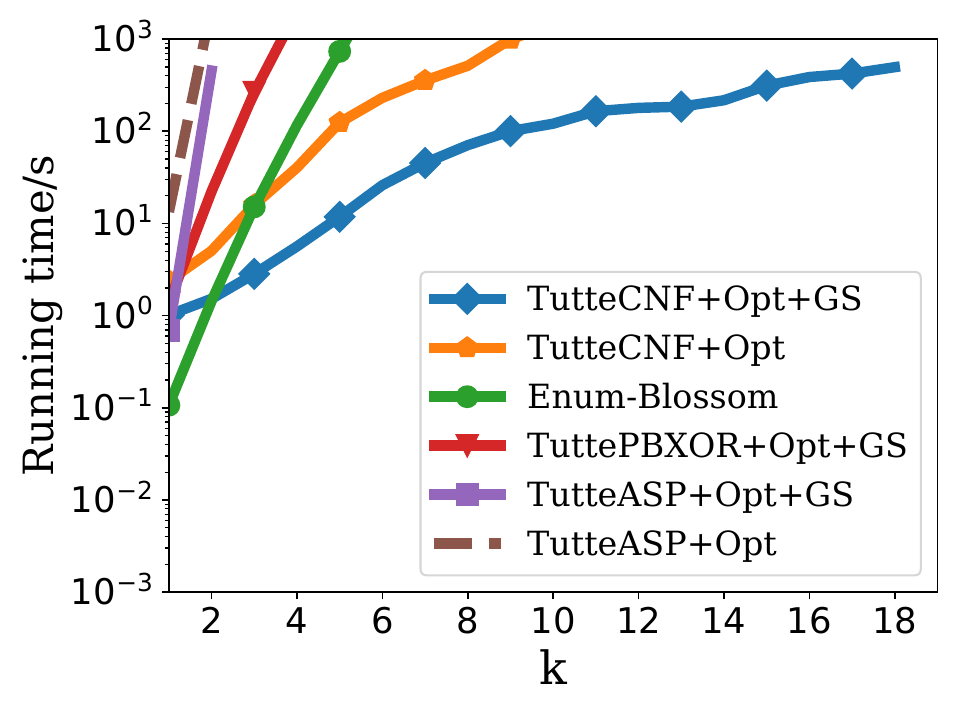}
\caption{Running time for verifying FORALL-PMVC condition 
 for graphs that satisfy \texttt{Dicke}$(36,k)$. QBF failed to solve a single instance even for $k=1$ within the time limit.}
\label{fig:dickenk}
\end{figure}

\paragraph{Experiment 2: Verifying graphs that satisfy the Dicke FORALL-PMVC conditions.}
 In this experiment, we use different techniques to verify graphs that satisfy the Dicke state. We generate two types of graphs. First, for $n\in \{6,8,\cdots,40\}$ we generate graphs that satisfy $\texttt{Dicke}(n,n/2)$ (see Section \ref{sec:quantum} and Figure \ref{fig:graphExamples}), where the size of legal coloring grows exponentially as $n$ increases.  Second, we fix $n=36$ and generate graphs that satisfy \texttt{Dicke}$(36,k)$ with $k\in \{1,2,\cdots, 18\}$. The Tutte-encoding-based approaches are supposed to output ``unsatisfiable", as the formula translates to finding a legal coloring with a Tutte set. The results are shown in Figure \ref{fig:dickenn2} and \ref{fig:dickenk}. Each running time is the median of $100$ independent executions. 

\begin{table}
		\centering
			\begin{footnotesize}
		\begin{tabular}{c c c}
			\toprule 
			Methods & & Running Time/s \\
			\cmidrule{1-1} \cmidrule{3-3}
			Blossom & & 0.02 \vspace{0.05cm}   \\ 
		ExactOne-PB & & 0.11 \vspace{0.05cm} \\ 
            TutteCNF+Opt+GS & & 10.61  \\
		TutteCNF+Opt & & 12.85  \\ 
  TutteASP+Opt+GP & & 14.89  \\
   TutteASP+Opt & & 66.40  \\ 
  TutteCNF & & 133.73  \\ 
  ExactOne-CNF & & Timeout ($>1000$s)  \\ 
			\bottomrule
		\end{tabular} 
			\caption{Running time for proving the uncolored complete bipartite graph $K_{48,50}$ has no perfect matching.}
		\label{table:bi}
			\end{footnotesize}
	\end{table}

\begin{figure}[t]
	\centering
\includegraphics[scale=0.46]{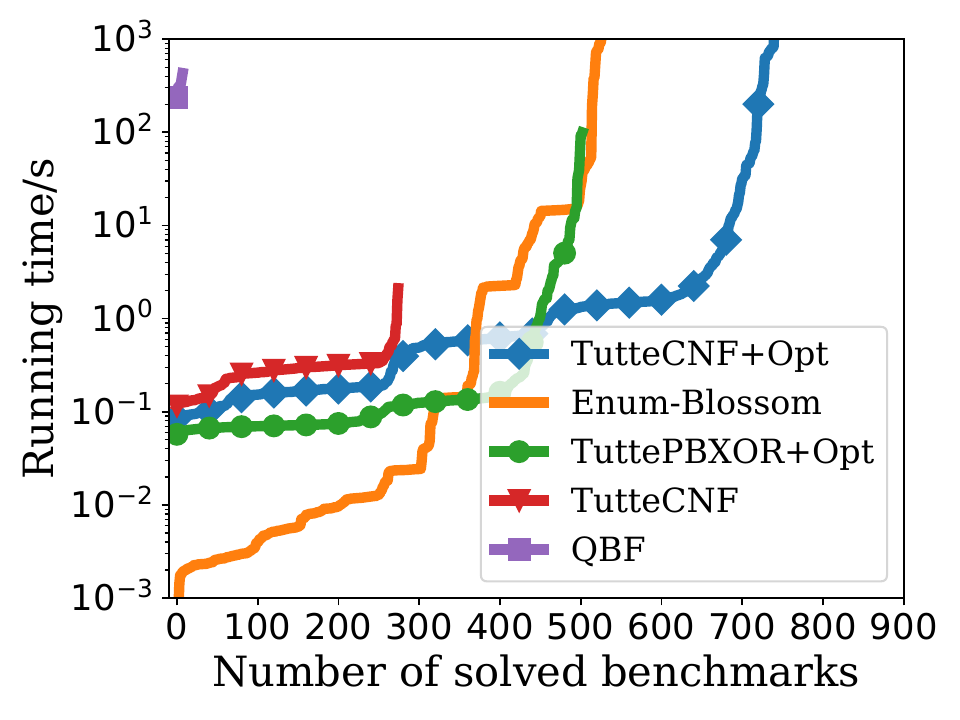}
\caption{The number of benchmarks solved by each method v.s. running time (1280 instances in total).}
\label{fig:sat}
\end{figure}

\paragraph{Experiment 3: Refuting graphs that violate the Dicke FORALL-PMVC conditions.}
 We evaluate the searching power of different approaches via refuting graphs 
 that violates the Dicke state. Tutte-encoding-based approaches are supposed to output ``satisfiable". Graphs in this benchmark are obtained by removing edges from graphs that satisfy the Dicke state. For each $(n,k)$ pair where $n\in \{10,15,20,\cdots, 80\}$, and $k\in\{0.1\cdot n,0.2\cdot n,0.3\cdot n, 0.4 \cdot n\}$, we first generate $20$ graphs that satisfy $\texttt{Dicke}(n,k)$. Then for each graph, we randomly remove either $40\%$ of blue edges or $2$ bicolored edges such that the resulting graph violates the Dicke state. In Enum-Blossom, we shuffled the list of legal colorings for more stable performance. The number of instances that each method solved v.s. the running time is plotted in Figure \ref{fig:sat}.

\subsection{Analysis of Results}
\noindent\textbf{Answer to RQ1:} 
A significant performance enhancement by optimization techniques in Section \ref{section:optimization} can be observed in Table \ref{table:bi} and Figure \ref{fig:dickenn2}-\ref{fig:sat}. For example, verifying \texttt{Dicke}$(10,5)$ was accelerated by $1200$ times from TutteCNF (122.7s) to TutteCNF+Opt (0.1s). On searching tasks (Figure \ref{fig:sat}), TutteCNF+Opt (741) solved 466 more instances than TutteCNF (275). Adding constraints for graph symmetry exploitation also yields great improvement for the verification of relatively large symmetric graphs. TutteCNF+Opt+GS successfully verifies all Dicke graphs in Experiment 2 within $1000$s, while TutteCNF+Opt timed out for verifying \texttt{Dicke}$(36,10)$ and \texttt{Dicke}$(34,17)$.  We conclude that our optimization techniques are critical for the performance of Tutte encoding. 

\noindent\textbf{Answer to RQ2:}  All experiments agree that CNF is the language leading to the best performance for Tutte encoding, with a significant advantage against PBXOR and ASP. The ASP solver Clingo is deployed with a relatively slow SAT solver Clasp with specialized adaptions, which makes it difficult to replace Clasp with a faster SAT solver, e.g., Kissat. To compare CNF encoding with ASP 
 fairly, in Experiment 2 we also solved the Tutte encoding by Clasp. In Figure \ref{fig:dickenn2}, TutteCNF+Opt (Clasp) still outperforms TutteASP+Opt+GS and TuttePBXOR+Opt+GS by far. We conclude that while high-level languages such as ASP and PBXOR can be convenient for modeling the problem, low-level languages such as CNF are necessary for better performance.

\noindent\textbf{Answer to RQ3:}
\paragraph{Experiment 1 (Table \ref{table:bi})}
When proving the non-existence of perfect matching in uncolored graphs $K_{n-2,n}$, the polynomial Blossom Algorithm is the fastest approach as expected.  Among all encoding-based approaches, \texttt{Exact-One} encoding with pseudo-Boolean solver (ExactOne-PB) scales best while ExactOne-CNF is the slowest. This observation aligns with the theoretical results about 1) the polynomial behavior of cutting-plane methods \cite{chandrasekaran2016cutting} and 2) the exponential lower bound of resolution-based approaches on perfect matching \cite{razborov2004resolution}. Compared with the Blossom Algorithm and ExactOnePB, approaches based on Tutte encoding take a longer time to solve the problem by searching for the unique Tutte set.

\paragraph{Experiment 2 (Figure \ref{fig:dickenn2}, \ref{fig:dickenk})} For the verification of Dicke FORALL-PMVC conditions, all approaches tend to scale exponentially. The QBF encoding presents the worst scalability, which failed to verify \texttt{Dicke}$(10,5)$ and \texttt{Dicke}$(36,1)$ within the time limit. Methods based on Vanilla Tutte encoding (TutteCNF) also do not scale well, which only outperform QBF. With optimization techniques, Tutte encodings in ASP and PBXOR languages (TuttePBXOR+Opt+GS, TutteASP+Opt+GS) were able to achieve better scalability, although the Enum-Blossom algorithm still performs better. Only when equipped with both optimization techniques and state-of-the-art SAT solvers (TutteCNF+Opt, TutteCNF+Opt+GS), does the Tutte encoding display promising potential. For verifying graphs that satisfy \texttt{Dicke}$(n,n/2)$, TutteCNF+Opt+GS can successfully verify graphs with $n=40$ and TutteCNF+Opt is able to solve for $n=34$ within $1000$s, both of which scale exponentially better than Enum-Blossom ($n=26$) and QBF ($n=8$) according to Figure \ref{fig:dickenn2}. As for verifying \texttt{Dicke}$(36,k)$ (Figure \ref{fig:dickenk}), when $k$ is small ($k\le 2$), the set of legal coloring is polynomially bounded and Enum-Blossom is the fastest algorithm. As $k$ increases, optimized CNF Tutte encoding with SAT solvers dominates other approaches by far.  

\paragraph{Experiment 3 (Figure \ref{fig:sat})}
On identifying graphs that violate FORALL-PMVC conditions, QBF only solved $8$ instances (1280 in total). TuttePBXOR+Opt and Enum-Blossom solved $503$ and $529$ instances respectively. Enum-Blossom is able to solve relatively small-size instances quickly (within $0.1$s). The optimized Tutte encoding in CNF (TutteCNF+Opt) solved $741$ instances, which outperforms Enum-Blossom mainly on large-size problems. 

\paragraph{Summary.}  Not surprisingly, the Blossom Algorithm dominates constraint-based approaches in proving the non-existence of PM for uncolored graphs. However, when solving FORALL-PMVC tasks with vertex coloring involved, our method displays promising performance, especially when the legal coloring set grows exponentially. The vanilla Tutte encoding does not scale well without optimization techniques and suitable language.  With modern CNF-SAT solvers, optimized Tutte encoding scales exponentially better than QBF and a graph theoretical algorithm (Enum-Blossom) on FORALL-PMVC problems. We conclude that while hybrid-constraint-based techniques are powerful and expressive, extra effort is necessary for solving FORALL-PMVC efficiently. Our approaches also scale significantly better than previous implementations for similar tasks from the quantum-computing community. The graph instances used in previous work have less than $15$ vertices \cite{cervera2021design}. 

\section{Conclusions and Future Directions} In this paper, we study a new application of hybrid Boolean constraints, named FORALL-PMVC, which arises in quantum computing. We propose a novel encoding based on Tutte's Theorem in graph theory as well as optimization techniques for this encoding. Empirical results demonstrate that in the most suitable encoding language (CNF), our approach based on optimized Tutte encoding scales significantly better than competing approaches on tasks with interest in quantum computing. Our study identifies the necessity of designing problem-specific encodings when applying powerful general-purpose constraint solvers. Our implementations also provide useful tools for the quantum computing community.

For future directions, we plan to apply constraint-based methods to solve problems beyond graph property checking. For example, the ultimate goal of this line of research  \cite{cervera2021design} is to \emph{construct} a graph w.r.t. a quantum state.  Considering an alternative constraint-programming approach for our problem based on filtering algorithms and \texttt{All-Different} constraints \cite{bessiere2010propagating} is another interesting direction.  It is also promising to apply our Tutte encoding for solving graph theoretical problems related to Tutte's Theorem algorithmically \cite{bauer2007tutte}.

\newpage

\section*{Acknowledgement}
Work supported in part by NSF grants IIS-1527668, CCF-1704883, IIS-1830549, DoD MURI grant N00014-20-1-2787, Andrew Ladd Graduate Fellowship of Rice Ken Kennedy Institute, and an award from the Maryland Procurement Office. We thank the anonymous reviewers for helping improve the paper.

\bibliographystyle{named}
\bibliography{ijcai22.bib}

\clearpage
\appendix
\section{Technical Proofs}
\subsection{Proof of Proposition \ref{prop:reducedGraph}}
$\Rightarrow:$ If there is a PM $P$ whose inherited vertex coloring is $c$, then every edge of $P$ must be in $E_c$. Therefore $G_c$ has a perfect matching.

\noindent $\Leftarrow:$ Suppose $G_c$ has a PM $P$. Then the color of every edge in $P$ agrees with $c$, which indicates $G$ has a perfect matching with inherited color $c$. 
\hfill\qedsymbol

\subsection{Proof of Proposition \ref{prop:tutteencoding}}
$\Rightarrow:$ Suppose the graph does not satisfy the FORALL-PMVC condition, we construct a solution to \texttt{TutteEncoding} as follows. Since the graph violates the FORALL-PMVC condition, there exists a legal coloring $c\in \mathcal{C}$ such that $G_c$ has no perfect matching. We first assign $\texttt{vc}_v^i$ to \texttt{True} if $c(v)=i$ and \texttt{False} otherwise. Then we have constraints in \texttt{ValidColoring} and \texttt{LegalColoring} satisfied. By Tutte's Theorem, $G_c$ has a Tutte set, say $S$. Then we assign $T_v=\texttt{True}$ if $v\in S$ and \texttt{False} otherwise. We assign $\texttt{e}_{uv}^{ab}$ to \texttt{True} if $\{(v,a),(u,b)\}$ is in $G[V\setminus S]_c$. The constraints in \texttt{RemainingEdges} are satisfied.  Suppose the non-empty components of $G[V\setminus S]_c$ has $k$ components and for each vertex $v\in V\setminus S$, let $\texttt{comp}(i)\in\{1,\cdots,k\}$ be the index of the connected  component that $v$ belongs to. We assign $\texttt{cc}_i^{\texttt{comp}(v)}$ to \texttt{True} and \texttt{cc}$_i^j$ to \texttt{False} for $j\neq \texttt{comp}(v)$, $j\in \{1,\cdots, |V|\}$. Then all constraints in \texttt{ConnectedComponents} are satisfied. Next, we assign $\texttt{Odd}_i$ to \texttt{True} if the $i$-th component has an odd number of vertices for ${i\in \{1,\cdots, |V|\}}$. Then we have constraints in \texttt{Odd} satisfied. Meanwhile, as each vertex $v\in V\setminus S$ is assigned with exactly one index of connected component, constraints in \texttt{ValidComponent} are also satisfied. Finally, by the definition of Tutte set, suppose $G[V\setminus S]$ has $\#odd$ connected components, then we have $\#odd > |S|$. Therefore \texttt{TutteCondition} is satisfied and \texttt{TutteEncoding} is finally satisfied.

    \noindent$\Leftarrow:$ Suppose \texttt{TutteEncoding} has a solution, we show that the graph does not satisfy the FORALL-PMVC condition. By Proposition \ref{prop:reducedGraph}, it is sufficient to prove that there exists a legal coloring $c\in \mathcal{C}$ such that $G_c$ has no perfect matching. We show that such a coloring is given by $c(v)=i$ if $\texttt{vc}_v^i=\texttt{True}$ for $v\in V$ in the solution. Since the solution satisfies \texttt{ValidColoring} and \texttt{LegalColoring}, the coloring $c$ is well-defined and legal. Applying Tutte's theorem, it is sufficient to show the existence of a Tutte set $S$ of $G_c$, which we construct as follows. For each $v\in V$, let $v\in S$ if $T_v=\texttt{True}$. Since \texttt{RemainingEdges} is satisfied, if $\texttt{e}_{uv}^{ab}$ is \texttt{True}, then $\{(u,a),(v,b)\}$ is in the edge set of $G[V\setminus S]_c$. Since \texttt{ConnectedComponent} is satisfied, if two vertices $u,v\in V\setminus S$ are in the same connected component, then vectors $(\texttt{cc}_u^1,\cdots,\texttt{cc}_u^{|V|})$ and $(\texttt{cc}_v^1,\cdots,\texttt{cc}_v^{|V|})$ must have same value. For each $i$ such that \texttt{Odd}$_i$ is \texttt{True}, let the vertex set $O_i\subseteq V\setminus S$ include all vertices $v\in V$ s.t. $\texttt{cc}_v^i=\texttt{True}$. We show that we can obtain an odd connected component from each $O_i$ for $i\in\{1,\cdots,|V|\}$ s.t. $\texttt{Odd}_i=\texttt{True}$. Since \texttt{Odd} is satisfied, $O_i$ has an odd number of vertices. Therefore, a connected component of $G[V\setminus S]$ in $O_i$ with an odd number of vertices must exist. Since \texttt{ValidComponent} is satisfied, all non-empty $\{O_i\}_{i\in\{1,\cdots,|V|\}}$ form a partition of vertices in $V\setminus S$. We can obtain at least $\sum_{i=1}^{|V|} \texttt{Odd}_i$ different odd connected components in $G[V\setminus S]_c$. By the \texttt{TutteCondition} we have $\sum_{i=1}^{|v|}\texttt{Odd}_i>\sum_{v\in V}T_v$. Therefore, there are more odd connected components in $G[V\setminus S]_c$ than vertices in $S$. By Tutte's Theorem, $G_c$ has no perfect matching and the FORALL-PMVC condition is violated by the graph.
   
\section{Generating Graphs w.r.t. Dicke State} In the following we show how to generate relatively sparse graphs that satisfy Dicke state. To generate \texttt{Dicke}$(n,k)$ (without loss of generality, suppose $k\le n/2$), we partition all $n$ vertices into $V_{1} = \{1,\cdots, k\}$ and $V_2=\{k+1,\cdots, n\}$. Then we add the following edges (we name Color $1$ as ``red" and $2$ as ``blue"):
\begin{itemize}
    \item For each $u\in V_1$ and $v\in V_2$, connect $u$ and $v$ with a red-blue edge and a blue-red edge.
    \item For each $u,v\in V_2$, connect $u$ and $v$ with a blue edge.
\end{itemize}

We call such a graph \texttt{DickeGraph}$(n,k)$. \texttt{DickeGraph}$(6,2)$ is shown in Figure \ref{fig:graph}. Next, we prove that such graphs satisfy the FORALL-PMVC condition for the Dicke state.
\begin{proposition}
    For each coloring with exactly $k$ red vertices and $n-k$ blue vertices, \texttt{DickeGraph}$(n,k)$ has a perfect matching with inherited vertex coloring $c$.
\end{proposition}
\begin{proof*}
     Suppose among the $k$ red vertices in $c$, $k_1$ of them are in $V_1$ and $k_2 = k-k_1$ of them are in $V_2$.  Then for the $k_1$ red vertices in $V_1$, match them with arbitrary $k_1$ blue vertices in $V_2$. For the $k_2$ red vertices in $V_2$, match them with $k_1$ blue vertices in $V_1$. For the rest $n-k-k_1-k_2=n-2k$ unmatched blue vertices in $V_2$, just match them with each other by blue edges (note that $n-2k$ is even). The inherited vertex coloring of this perfect matching has exactly $k$ red vertices.  
\end{proof*}
Moreover, \texttt{DickeGraph} has no illegal perfect matchings.
\begin{proposition}
    All perfect matchings of \texttt{DickeGraph}$(n,k)$ are legal w.r.t. Dicke State.
\end{proposition}
\begin{proof*}
    Since vertices in $V_1$ can be only matched with vertices in $V_2$ via bi-colored edges, a perfect matching of \texttt{DickeGraph} contains exactly $k$ blue-red edges and $n-k$ blue edges. Therefore all perfect matchings have legal colorings.
\end{proof*}

Some bi-colored edges in \texttt{DickeGraph} are necessary, as formalized in the following proposition.
\begin{proposition}
    Remove an edge $\{(u,blue),(v,red)\}$ s.t. $v\in V_1, u\in V_2$ from \texttt{DickeGraph}$(n,k)$ yields to a graph that violates the FORALL-PMVC condition.
\end{proposition}
\begin{proof*}
  We construct a legal vertex coloring $c$ s.t. $G_c$ has no perfect matching as follows. Let $u$ be colored red and all other vertices in $V_2$ be colored blue. Let $v$ be colored blue and all other vertices in $V_1$ be colored red. It is not hard to verify that both $v$ and $u$ are isolated in $G_c$. Therefore $G_c$ has no perfect matching.  
\end{proof*}

Removing some blue edges that connect vertices in $V_2$ does not necessarily make the graph violate the Dicke FORALL-PMVC condition. For example, all blue edges of \texttt{DickeGraph}$(n,n/2)$ can be removed while the Dicke FORALL-PMVC condition is still satisfied.

\section{Comprehensive Experiment Results of Experiment 1} 
\begin{figure}[t]
	\centering
\includegraphics[scale=0.52]{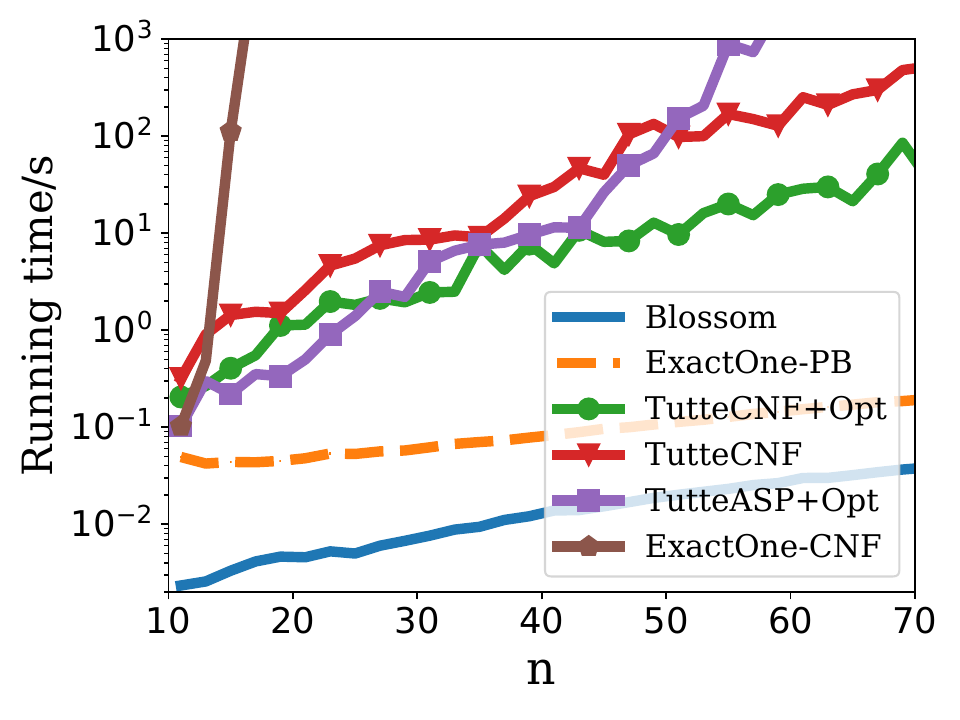}
\caption{Running time for proving the uncolored complete bipartite graph $K_{n-2,n}$ has no perfect matching.} 
\label{fig:bi}
\end{figure}

More results of proving the uncolored complete bipartite graph $K_{n-2,n}$ has no perfect matching (Experiment 1) are shown in Figure \ref{fig:bi}. As for checking the existence of perfect matchings, Blossom Algorithm is known to scale polynomially and it outperforms other algorithms in the experiment. Among constrained-based approaches, ExactOne-PB has the best performance while ExactOne-CNF scales worst. Approaches based on Tutte encoding scale exponentially on those tasks, while optimization techniques still improve the performance.
\end{document}